\documentclass[aos]{imsart}

\RequirePackage{amsthm,amsmath,amsfonts,amssymb}
\RequirePackage[numbers,sort&compress]{natbib}
\RequirePackage[colorlinks,citecolor=blue,urlcolor=blue]{hyperref}
\RequirePackage{graphicx}

\usepackage{graphicx}
\usepackage{caption}
\usepackage{subcaption}
\usepackage{float}
\usepackage{hyperref}

\startlocaldefs
\theoremstyle{plain}

\newtheorem{theorem}{Theorem}[section]
\newtheorem{lemma}[theorem]{Lemma}
\newtheorem{corollary}{Corollary}
\theoremstyle{remark}
\newtheorem{definition}[theorem]{Definition}

\endlocaldefs

\usepackage{algorithm}
\usepackage{algorithmic}

\newcommand{\arms}{K}
\newcommand{\K}{\arms}
\newcommand{\selectedarm}{A_t}
\newcommand{\reward}{X_t}

\newcommand{\totalsteps}{T}

\newcommand{\regret}{\mathcal{R}_T}
\newcommand{\meanmax}{\mu^\star}
\newcommand{\Prob}{\mathbb{P}}

\newcommand{\Ncal}{\mathcal{N}}

\newcommand{\lastpull}[1]{T_{#1}}
\newcommand{\numpull}{T}
\newcommand{\T}{T}
\newcommand{\N}{N}
\newcommand{\xbar}{\bar{X}}
\newcommand{\bonus}[2]{\sqrt{\frac{2 \log \numpull}{n_{#1, #2}}}}

\newcommand{\lilbound}[1]{\frac{\gT}{\sqrt{#1}}}
\newcommand{\gT}{g_{\T}}
\newcommand{\Exs}{\mathbf{E}}

\newcommand{\nopt}{n^\star}
\newcommand{\Tprime}{T^\prime_\epsilon}
\newcommand{\goodevent}{\ensuremath{\mathcal{E}}}
\newcommand{\dir}{u}
\newcommand{\Cint}{\mathcal{C}}
\newcommand{\RR}{\mathbb{R}}

\newcommand{\sigmahat}{\hat{\sigma}}
\newcommand{\subG}{\lambda}
\newcommand{\rvar}{\sigma}
\newcommand{\rvarhat}{\hat{\rvar}}
\newcommand{\B}{B}
\newcommand{\Barms}{\mathcal{S}_{\B}}
\newcommand{\filt}[1]{\mathcal{F}_{#1}}
\newcommand{\real}{\mathbb{R}}
\newcommand{\armdist}{\mathcal{P}}

\graphicspath{{../Code/Paper-code/paper-plots/}}

\begin{document}

\begin{frontmatter}
\title{Inference with the Upper Confidence bound Algorithm}
\runtitle{Inference with UCB}

\begin{aug}
\author[A]{\fnms{Koulik}~\snm{Khamaru}\ead[label=e1]{kk1241@stat.rutgers.edu}}
\and
\author[B]{\fnms{Cun-Hui}~\snm{Zhang}\ead[label=e2]{czhang@stat.rutgers.edu}}
\address[A]{Department of Statistics,
Rutgers University\printead[presep={,\ }]{e1}}
\address[B]{Department of Statistics,
Rutgers University\printead[presep={,\ }]{e2}}
\end{aug}

\begin{abstract}
In this paper, we discuss the asymptotic behavior of the Upper Confidence Bound (UCB) algorithm in the context of multiarmed bandit problems and discuss its implication in downstream inferential tasks. While inferential tasks become challenging when data is collected in a sequential manner, we argue that this problem can be alleviated when the sequential algorithm at hand satisfy certain stability properties. 
This notion of stability is motivated from the seminal work of Lai and Wei~\cite{lai1982least}.  Our first main result shows that such a stability property is always satisfied for the UCB algorithm, and as a result the sample means for each arm are asymptotically normal. Next, we examine the stability properties of the UCB algorithm when the number of arms $K$ is allowed to grow with the number of arm pulls $\T$. We show that in such a case the arms are stable when $\frac{\log K}{\log \T} \rightarrow 0$, and the number of \emph{near-optimal} arms are large. 
\end{abstract}



\end{frontmatter}

\newcommand{\rr}[1]{\textcolor{red}{#1}}

\section{Introduction}
\label{sec:Intro}
Reinforcement learning (RL) has emerged as a cornerstone of artificial intelligence, driving breakthroughs in areas from game-playing agents to robotic control. Its ability to learn optimal decision-making strategies through environmental interaction has positioned RL as a key technology in the development of autonomous systems. Central to RL is the exploration-exploitation dilemma, where agents must balance discovering new information with leveraging known high-reward options.  The Upper Confidence Bound (UCB) algorithm addresses this dilemma through the principle of optimism in the face of uncertainty. By maintaining upper confidence bounds on the expected rewards of each action, UCB provides a theoretically grounded approach to balancing exploration and exploitation in various RL settings.

However, the adaptive nature of data collection in RL violates the independent and identically distributed (i.i.d.) assumption underpinning many statistical methods. This sequential dependency poses significant challenges for analysis and inference in RL contexts. Despite these challenges, robust statistical inference remains crucial for RL. It enables quantification of uncertainty, validation of model performance, and provision of reliability guarantees—essential factors as RL systems are deployed in increasingly critical applications.

This paper investigates a novel stability property in adaptive RL algorithms. We show that the UCB algorithm induces a form of \emph{stability} in the sequential non-iid data which makes the downstream statistical inference process straightforward. For instance, under this stability property classical statistical estimators asymptotically normal.

\subsection{Related Work}
\label{sec:related-work}
In this section, we provide a brief survey existing literature on Multiarmed bandits, the UCB algorithm, and inference with data generated from sequential procedures.

\subsubsection{Multiarmed bandits and the UCB algorithm}
The study of multi-armed bandits has a rich history, dating back to the seminal work of Thompson~\cite{thompson1933likelihood} and Robbins~\cite{robbins1952some}. The Upper Confidence Algorithm (UCB) algorithm, introduced by Lai and Robbins~\cite{lai1985asymptotically} and Lai~\cite{lai1987adaptive}  and later refined by Auer~\cite{auer2002using}, has become a cornerstone of the field due to its strong theoretical guarantees and empirical performance. The early (regret) analysis of the UCB algorithm is by Kathehakis and Robbins~\cite{katehakis1995sequential} and later refined by~\cite{agrawal1995sample}. The UCB algorithm is based on an principle of \emph{optimism in the face of uncertainty}, which turned out to be very influential in the area of reinforcement learning and sequential decision-making in general. We refer the reader to the book~\cite{lattimore2020bandit} and the references therein for a discussion and several applications of this principle.

\subsubsection{Statistical inference with adaptively collected data}
While sequential decision-making procedures like the UCB are known to yield low regret, the sequential nature of the algorithm induces a 
dependence in the resulting data-set. Stated differently, the data collected by sequential data generating processes like UCB are not iid, thus it is not clear how can we perform downstream statistical inference using the adaptively collected data-set. 

The challenge of performing statistical inference with adaptively collected data has been recognized in the statistical literature for at least 40 years. In the context of time series analysis, works by Dickey and Fuller~\cite{dickey1979distribution}, White~\cite{white1958limiting,white1959limiting}, and Lai and Wei~\cite{lai1982least} highlighted the breakdown of classical asymptotic theory when the data is generated from an regressive time series model. More recent work~\cite{deshpande2018accurate,deshpande2023online,khamaru2021near,lin2023semi,lin2024statistical,ying2024adaptive} highlighted, via numerical simulations, that a similar phenomenon can occur in bandit problems. Zhang et al.~\cite{zhang2020inference} proved that in the case of batched bandits Z-estimators (sample arm means for multi-armed bandits) may not be  asymptotic normal when the data is collected via popular bandit algorithms like UCB, Thompson sampling, and the $\epsilon$-greedy algorithm.

In order to perform valid statistical inference, researchers have suggested two types of approaches. The first approach is non-asymptotic in nature, and is based on the concentration bounds on self-normalized Martingales; see the works~\cite{abbasi2011improved,shin2019bias,waudby2023anytime} and  the references therein.  
These works stem from the seminal works of~\cite{de2004self,pena2009self}, 
and provide confidence intervals that are valid for any sample size and are usually more conservative (larger confidence intervals). The second approach is asymptotic in nature, and it exploits the inherent Martingale nature present in data via Martingale central limit theorem and debiasing~\cite{hall2014martingale,zhang2014confidence}. The confidence intervals obtained are valid asymptotically, and are often shorter than the ones obtained from concentration inequality based methods. See the works~\cite{deshpande2018accurate,deshpande2023online,khamaru2021near,lin2023semi,lin2024statistical,ying2024adaptive,zhang2020inference,hadad2021confidence,bibaut2021post,zhan2021off,syrgkanis2023post} and the references therein for an application of this technique. 

\subsection{Failure of classical estimators: A closer look}
\label{sec:A-closer-look}
While classical guarantees are not valid under when the data is collected via sequential methods, it is worthwhile to understand whether that is always the case. To motivate this discussion, we consider two algorithms for collecting data in a 2-armed bandit problem with arm means vector $(\mu_1, \mu_2) \equiv (0.3, 0.3)^\top$ with standard Gaussian error. 
\begin{enumerate}
    \item[(a)]  The arms are selected using an $\epsilon$-greedy algorithm with $\epsilon = 0. 1$. 
    \item[(b)] The arms are selected using the UCB algorithm, detailed in Algorithm~\ref{algo:UCB}.
\end{enumerate}

\begin{figure}[htbp]
    \centering
    \begin{subfigure}[b]{0.49\textwidth}
        \centering
        \includegraphics[width=\textwidth]{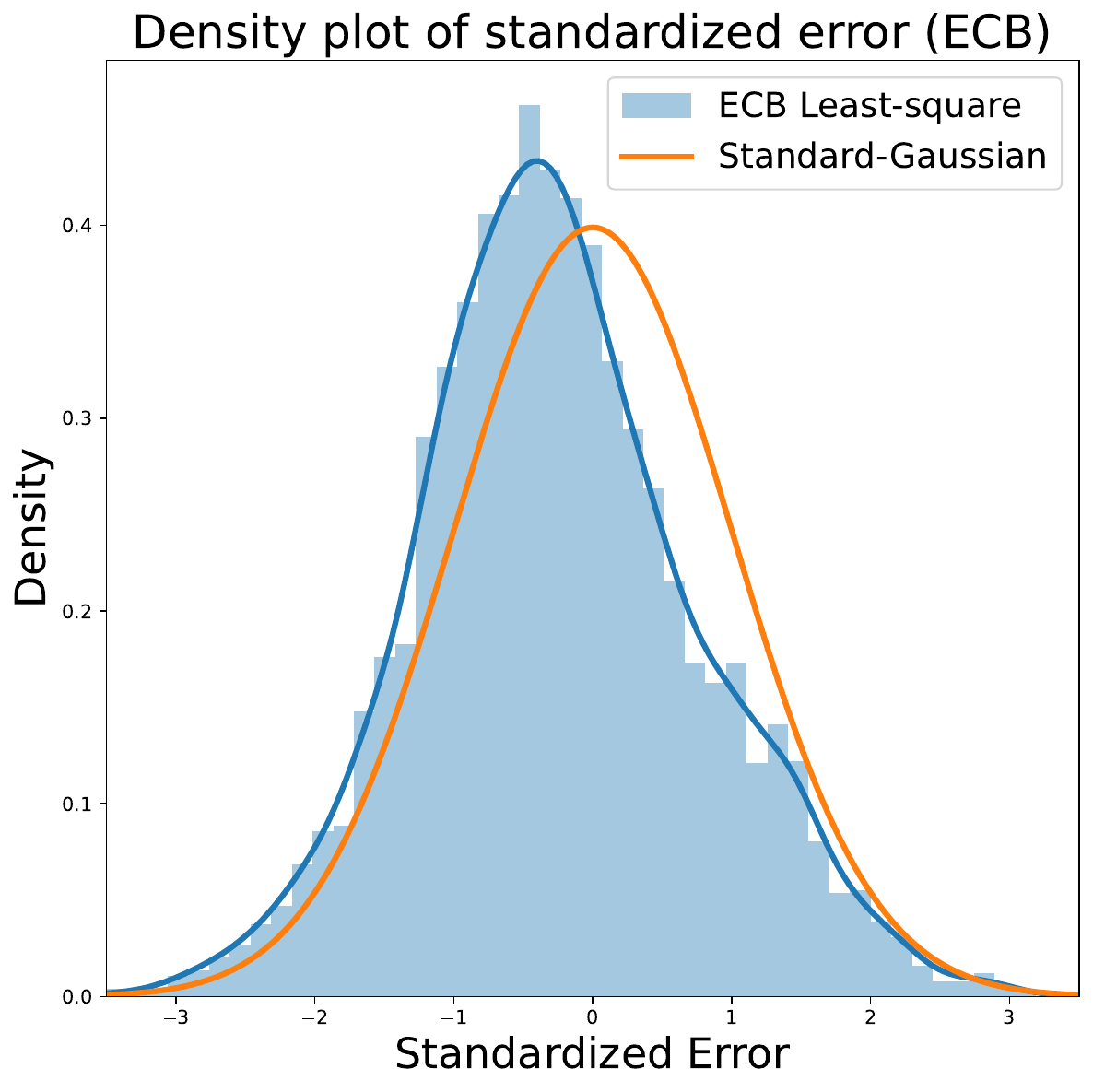}
        \caption{$\epsilon$-greedy algorithm.}
    \end{subfigure}
    \hfill 
    \begin{subfigure}[b]{0.49\textwidth}
        \centering
        \includegraphics[width=\textwidth]{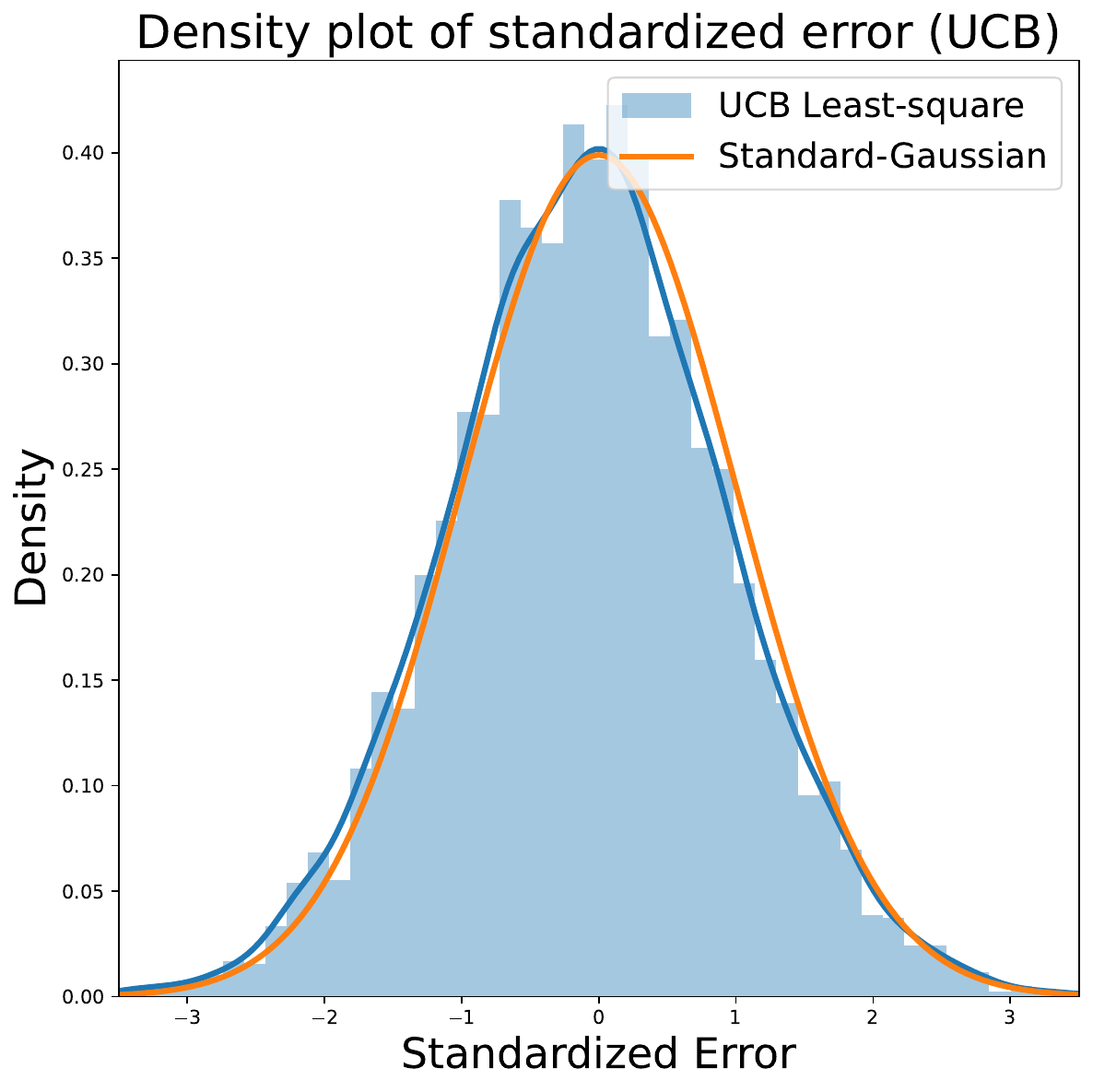}
        \caption{UCB algorithm.}
    \end{subfigure}
    \caption{Distribution of $\frac{\hat{\mu}_2 - \mu_2}{\sqrt{N_{2}}}$, the standard error  of sample arm mean for arm 2. The distribution of the sample mean deviates from standards normal distribution (left panel) when the $\epsilon$-greedy algorithm is used. The distribution is in good accordance with a standard normal distribution when the UCB algorithm is used (right panel).
    The results are averaged over $5000$ repetitions. See Section~\ref{sec:A-closer-look} for a detailed discussion. The code for the plot can be found in  this~\href{https://github.com/KoulikBerkeley/UCB-stability-plots/tree/main}{Github-repo}.} 
    \label{fig:greedy-vs-ucb}
\end{figure}

Figure~\ref{fig:greedy-vs-ucb} plots the asymptotic distribution of $\frac{\hat{\mu}_2 - \mu_2}{\sqrt{N_{2}}}$, the standard error of sample arm mean for the second arm, for both the algorithm. The results are averages over 5000 simulations. The figure reveals that the while the sample arm mean is \emph{not asymptotically normal} when the $\epsilon$-greedy algorithm is used, and is the distribution is asymptotically normal when the UCB algorithm is used; we provide a formal proof of this asymptotic normality in Theorem~\ref{thm:CI}. Put simply, the failure of classical method is dependent on the sequential method used. Thus, it is natural to ask: 
\begin{center}
\emph{Which sequential methods / algorithms preserve the asymptotic normality \\ properties of classical statistical methods?}   
\end{center}
This problem was partly discussed in the seminal work of Lai and Wei~\cite{lai1982least} in the context of stochastic linear model:
\begin{align*}
y_i = \left\langle x_i, \theta^\star \right\rangle + \epsilon_i
\end{align*}
where $x_i \in \sigma\left(\{x_1, y_1, \ldots, x_{i - 1}, y_{i - 1} \}\right)$ --- the $\sigma$-field generated by the data collected up to round $i - 1$, and $\epsilon_i's$ are iid noise. Clearly, the multi-armed bandit problem is a special case of this model. 
Lai and Wei argued that in the context of linear model the ordinary least square estimator is asymptotically normal if the sample \emph{covariance matrix stabilizes}. Concretely, there exists a sequence of deterministic positive definite matrices $\{B_n\}_{n \geq 1}$
\begin{align*}
    B_n^{-1} S_n \stackrel{p}{\rightarrow} \mathbf{I}  
\end{align*}
where $S_n = \sum_{i = 1}^n x_i x_i^\top$  is the sample covariance matrix. In the context of multi-armed bandit problems the above result translates to the stability of the arm-pulls. Concretely, the  
number of times any arm is pulled converges to a non-random limit when the total number of arm pulls grows to infinity.  (UCB) algorithm.
\subsection{Contributions}
\label{sec:Contribution}
The main contribution of this paper is to show that the such a \emph{stability}
property is satisfied for when we use the upper confidence bound (UCB) algorithm in a $K$-armed bandit problem. Specifically, we prove
\begin{itemize}
\item[(a)] The number of times an arm $a \in [K]$ is pulled by the UCB algorithm converges in probability to a deterministic limit. 
\item[(b)] We argue that this stability property enables us to perform statistical inference on the data collected by the UCB algorithm as if the dataset were independent and identically distributed (i.i.d.). For instance the sample-mean for any arm is an asymptotically normal estimator for the true arm mean.
\item[(c)] Finally, we show that such stability property is also satisfied when the number of arms are allowed to grow with the number of rounds. 
\end{itemize}
\subsection{Organization:}
The remainder of this paper is organized as follows: Section~\ref{sec:Set-up} provides background on the multi-armed bandit problem and the UCB algorithm. In Section~\ref{sec:Main} we introduce our main stability result
and discuss its consequences in the context of statistical inference. In Section~\ref{sec:large-arms}, we discuss how these stability properties in bandits can be extended to the case where the number of arms, denoted by  $K$, is allowed to grow with $T$, the number of samples (arm-pulls). Section~\ref{sec:Main-proofs}  is devoted to the proofs our main results. 
Finally, in Section~\ref{sec:Discussion} we discuss the broader impact of our work and suggest future directions. 

\section{Problem set up}
\label{sec:Set-up}
In this paper, we consider a multiarmed bandit problem with $\K$ arms. At each time step $t$, the player selects an arm $\selectedarm \in [\K]$ and receives a random reward $\reward \in \real$ from the distribution $\armdist_{A_t}$. 
Let $\mu_a$ and $\sigma_a^2$, respectively, denote the mean and variance of the distribution $\armdist_a$,  
and 
$\mu = (\mu_1, \ldots, \mu_K)^\top$  
denote the vector of arm-means. The goal of bandit algorithms is to 
maximize the total expected reward or equivalently to minimize the regret, defined as the difference between the total expected reward obtained and the maximum possible total reward, over $\totalsteps$ time steps:
\begin{align}
\label{eq:minimize_regret}
\texttt{(Regret:)} \qquad  \regret =   \totalsteps \cdot \meanmax  - \Exs[\sum_{t=1}^{\totalsteps} \reward]  \qquad \qquad \text{where} \quad \mu^\star = \max_{k \in [\K]} \mu_k.
\end{align}
 
In this paper, we are interested in the behavior of the celebrated Upper Confidence Bound algorithm~\cite{lai1985asymptotically,lai1987adaptive,auer2002using}. We detail the algorithm in Algorithm~\ref{algo:UCB}. 
Throughout we assume that $\T$ is known to the algorithm. 
\begin{algorithm}
\caption{UCB algorithm}
\label{algo:UCB}
\begin{algorithmic}[1]
\STATE Pull once each of the $K$ arms 
in the first $K$ iterations. 
\FOR{$t = K + 1, \ldots, \T - 1$}
    \STATE Compute the UCB boundary
\begin{align}
\label{eqn:ucb-rule-mod}
\mathrm{UCB}(a,t) := \xbar_{a,t} + \sqrt{\frac{2\log \T}{n_{a, t}}}
\end{align}
\STATE Choose arm $A_t$ given by 
\begin{align}
\label{eqn:arm-choose-rule-ucb}
A_{t + 1} = \arg\max_a \;\; \mathrm{UCB}(a,t)
\end{align}
\ENDFOR
\end{algorithmic}
\end{algorithm}

\vspace{5pt}

\noindent
It is well-know that the sequential nature of the bandit algorithms makes the downstream inference process challenging~\cite{deshpande2018accurate,deshpande2023online,khamaru2021near,lin2023semi,ying2024adaptive,zhang2020inference}. We would like to understand whether this is the case for the Upper Confidence Bound algorithm~\ref{algo:UCB}.  Concretely, let $n_{a,\T}$ and $\bar{X}_{a, \T} :=
(\sum_{t = 1}^{\T} X_t \cdot 1_{A_t = a})/n_{a, \T}$ denote respectively the sample size and sample mean of the rewards associated with arm $a$. We would like to identify conditions under which $\xbar_{a,\T}$ is asymptotically normal:
\begin{align}
\label{eqn:normality-condition}
\hspace*{-4cm}
\text{(Normality:)}
\qquad 
\sqrt{n_{a, \T}} \cdot \left( \bar{X}_{a, \T} - \mu_a \right) \stackrel{d}{\rightarrow} \mathcal{N}(0, \sigma_a^2)
\end{align}

\subsection{Asymptotic normality and arm-pull stability}
We now connect the asymptotic normality property~\eqref{eqn:normality-condition} of arm $a$ with that of certain stability property of $n_{a,\T}$, the number of times arm $a$ is pulled in $\T$ rounds. 
Note that $n_{a, \T}$ is random. 

\newcommand{\naopt}{n^\star_a}

\begin{definition}
An arm $a \in [K]$ is said to be stable if there exists non-random scalar $n^\star_a$ such that 
\begin{align}
\label{eqn:stability}
\hspace{-4cm} \text{(Stability:)} \qquad \;\;\;\;
\frac{n_{a, \T}}{\naopt} \stackrel{p}{\rightarrow} 1 \quad \text{with} \quad \nopt_a \rightarrow \infty
\end{align}
Here, the scalar $\naopt$ is allowed to depend on $\T, \{\mu_a\}_{a \in [K]}, \{\sigma^2_a\}_{a \in [K]}$. 
\end{definition}
This dentition is motivated from the seminal work of Lai and Wei~\cite{lai1982least}, where the authors used a similar condition to prove asymptotic normality of the least square estimators in a stochastic regression model. We now show that if arm $a$ is stable, then the asymptotic 
normality~\eqref{eqn:normality-condition} holds for arm $a$.

\newcommand{\Cov}{\textrm{Cov}}

Let $\filt{t - 1}$ denote the $\sigma$-filed generated by $\{X_1, \ldots, X_{t - 1}\}$. We assume that the arm reward distributions $\{\armdist_a\}$ are $1$-sub-Gaussian, and the number of arms $K$ is fixed for simplicity (we relax this assumption at a later section). By definition~\eqref{eqn:arm-choose-rule-ucb} we have that $A_t \in \filt{t - 1}$, and  
given an arm $a$, the sum $ \frac{1}{\sigma_a\sqrt{\nopt_a(\T)}} \sum_{t = 1}^\T  1_{A_t = a} \cdot \left(X_t - \mu_a\right) $ is a sum of Martingale difference sequence. We have 
\begin{align*}
\sum_{t =1}^{\T} \text{Var}\left( \frac{1}{\sigma_a\sqrt{\nopt_a(\T)}} 1_{A_t = a} \cdot \left(X_t - \mu_a\right) \Bigg| \filt{t - 1} \right) = \frac{n_{a, \T}}{\nopt_a(\T)} \stackrel{p}{\rightarrow} 1. 
\end{align*}
Stated differently, the sum of the conditional variances of the Martingale difference array stabilizes. Additionally, using the assumption $\nopt_a \rightarrow \infty$ and  sub-Gaussian property of the reward distribution we have that the Lindeberg condition of Triangular array is satisfied. Applying the Martingale CLT for triangular array \cite{dvoretzky1972asymptotic, hall2014martingale}, and using Slutsky's theorem we conclude
\begin{align} 
\label{eqn:asymptotic-normality-arm-a}
\frac{\sqrt{n_{a, \T}}}{\sigmahat_a} \cdot \left( \bar{X}_{a, \T} - \mu_a \right) \stackrel{d}{\rightarrow} \mathcal{N}(0, 1)
\end{align}
where $\hat{\sigma}^2_a$ is a consistent estimate of the variance $\sigma_a^2$.

Put simply, whenever the \emph{stability} condition is satisfied, the arm means $\bar{X}_{a, \T}$ are \emph{asymptotic normal} and we can construct asymptotically exact $1 - \alpha$ confidence interval for the mean $\mu_a$. In the next section, where we detail our main results, we discuss conditions under which the stability condition~\eqref{eqn:stability} holds for the Upper Confidence Algorithm~\ref{algo:UCB}.

\section{Main results}
\label{sec:Main}
Here we provide our main result which shows stability of the UCB algorithm. Without loss of generality we assume that arm $1$ is among optimal arms, and we do not assume that the optimal arm is unique. We also assume that 
\begin{subequations}
\begin{align}
\label{eqn:arm-mean-diff-UB}
& 0 \leq \frac{\Delta_a}{\sqrt{2 \log \T}} = o(1) 
\quad \text{for all arms a}. \\
 & \mathcal{P}_a \;\; \text{is} \;\; \subG_a-\text{sub-Gaussian} \;\; \text{for all arms a, and} \;\; |\subG_a| \leq B \;\; \text{for some constant} \;\; B < \infty.
 \label{eqn:sub-Gaussian}
\end{align}
\end{subequations}
We will explain the significance of these assumptions shortly. Our main result is the following:  

\begin{theorem}
\label{thm:bandit-stability}
Suppose we pull bandit arms using Algorithm~\ref{algo:UCB}. Let Assumptions~\eqref{eqn:arm-mean-diff-UB}-\eqref{eqn:sub-Gaussian} be in force and the number of arms $K$ fixed. Then, for each arm $a \in [K]$, the number of arm pulls 
$n_{a, \T}$ satisfies
\begin{align}
\label{eqn:arm-a-stability}
\frac{n_{a, T}}{\left( 1/\sqrt{\nopt}
+ \sqrt{\Delta_a^2 / 2 \log \T} \right)^{-2}} \rightarrow 1 \qquad \text{in probability}.
\end{align}
where $\nopt \equiv \nopt(\T, \{\Delta_a\}_{a \in [K]})$ is the unique solution to the following equation
\begin{align}
\label{eqn:n-star-eqn}
\sum_{a } 
\frac{1}{\left(\sqrt{\T/\nopt}
+ \sqrt{\T \Delta_a^2 / 2 \log \T} \right)^{2}} = 1
\end{align}
Here, $\Delta_a = \mu_1 - \mu_a$ and without loss of generality we assumed that arm 1 is among the optimal arms.  
\end{theorem}

A few comments are on the order. 
Condition~\eqref{eqn:arm-mean-diff-UB} ensures that arm $a$ is pulled infinitely often. Concretely, 
it ensures that $n_{a, \T} \rightarrow \infty$ in probability. We refer the reader to the seminal works~\cite{lai1985asymptotically,lai1987adaptive} for a similar condition.   

It is worthwhile to understand the consequence of Theorem~\ref{thm:bandit-stability} in some special cases. Suppose $\Delta_a = o(\sqrt{\frac{\log \T}{\T}})$ for some arm $a$. Then~\eqref{eqn:arm-a-stability} ensures that 
\begin{align*}
\frac{n_{a, \T}}{\nopt} \rightarrow 1 \qquad \text{in probability whenever} \quad \Delta_a = o \left( \sqrt{\frac{\log \T}{\T}} \right).
\end{align*}
Put simply, near-optimal arms are pulled equally often asymptotically.

\subsection*{Comparison to prior work} It is interesting to compare Theorem~\ref{thm:bandit-stability} with an earlier work of~\cite{kalvit2021closer}. 
Parts (II) and (III) of Theorem 1 in~\cite{kalvit2021closer} provide the limiting distribution of the number of arm pulls when $\Delta = \mu_1 - \mu_2 \leq \sqrt{\frac{\theta \log \T}{\T}}$ for some $\theta > 0$. Theorem~\ref{thm:bandit-stability} recovers these results when we substitute $K = 2$.
 When $\frac{\Delta}{\sqrt{\frac{\log \T}{\T}}} \rightarrow \infty$, Part (I) of Theorem~1 in \cite{kalvit2021closer} provides the limiting distribution of $n_{1, \T}$. On the contrary, Theorem~\ref{thm:bandit-stability} characterizes the behavior of both $n_{1, \T}$ and $n_{2, \T}$. Finally, Theorem~\ref{thm:bandit-stability} provides the limiting distribution of $K$ arms for any choices of $\{\Delta_a\}$, thereby resolving the problem that was left open in~\cite{kalvit2021closer}.

\subsection{Statistical inference:}
Theorem~\ref{thm:bandit-stability} allows us to provide an asymptotically exact $1 - \alpha$ confidence interval. Let the $\rvar^2_a$ denote the variance of the rewards associated with arm $a$.  Given a fixed direction $\dir = (\dir_1, \ldots, \dir_K)^\top \in \RR^{\K}$, define 
\begin{align}
\label{eqn:CI-normality}
\Cint_{\dir, \alpha} = \left[ \dir^\top \xbar_{\T} -  z_{1 - \alpha/2} \cdot \sum_{a = 1}^{\K} \frac{\rvarhat_a \dir_a^2}{\N_{a,\T}}, \qquad \dir^\top \xbar_{\T} +  z_{1 - \alpha/2} \cdot  \sum_{a = 1}^{\K} \frac{ \rvarhat_a \dir_a^2}{\N_{a,\T}}  \right]
\end{align} 
where $\rvarhat_a$ is any consistent estimate of $\rvar_a$, and $z_{1 - \alpha/2}$ is the $1 - \alpha/2$ quantile of the standard normal distribution, and $\xbar_{\T} = (\xbar_{1, \T} \ldots, \xbar_{K, \T})^\top$.
\begin{theorem}
\label{thm:CI}
Suppose the conditions of Theorem~\ref{thm:bandit-stability} are in force. Then, 
given a fixed direction $\dir \in \RR^{\K}$ and $\alpha \in (0,1)$, the confidence interval $\Cint_{\dir, \alpha}$ defined in~\eqref{eqn:CI-normality}  satisfies
\begin{align*}
\lim_{\T \rightarrow \infty} 
\Prob(\Cint_{\dir, \alpha} \ni \dir^\top \mu) = 1 - \alpha.
\end{align*} 
\end{theorem}

\subsubsection*{Consistent estimator of variance}
The last theorem allows us to construct asymptotically exact confidence interval for $\dir^\top \mu$ given we have consistent estimators of variances $\sigma^2_a$. For arm $a \in [K]$ define 
\begin{align}
\label{eqn:sigmahat-defn}
\sigmahat^2_a := \frac{1}{n_{a, \T}} \sum_{t = 1}^{\T} \left( X_t - \xbar_{a, \T} \right)^2\cdot 1_{A_t = a}
\qquad \text{with} \quad  \xbar_{a, \T} = \frac{1}{n_{a, \T}} \sum_{t = 1}^{\T} X_{a, t} \cdot 1_{A_t = a}.
\end{align}
Then we have 
\begin{corollary}
\label{cor:var-estimate}
Let the assumptions of Theorem~\ref{thm:bandit-stability} are in force. Then for all arm ${a \in [K]}$, the variance estimator $\sigmahat_a^2$ from~\eqref{eqn:sigmahat-defn} is consistent for $\sigma_a^2$.
\end{corollary}
 See Section~\ref{proof-cor-uneq-var} for a proof of this corollary. 
\section{Can we let the number of arms grow?}
\label{sec:large-arms}

In this section, we study the stability properties of arms when the number of arms grows with the number of arm pulls $\T$. Unfortunately, some of the proof techniques used in the previous section does not apply when the number of arms $K = K(\T)$ is allowed to grow with the number of arm pulls $\T$. The fist assumption that we need is 
\begin{align}
\label{eqn:K-growth}
\frac{\log K}{\log \T} \rightarrow 0 \qquad \text{as} \;\; \T \rightarrow \infty.  
\end{align}  
In other words, $K$ grows slower than any positive power of $\T$. This assumption ensures that a finite sample version of the law of iterated logarithm holds simultaneously for all $K$ arms. We start with the definition of near-optimal arms. Again, without loss of generality 
we assume that arm $1$ is among the optimal arms.

\subsection*{Near optimal arms:} Given a constant $\B > 0$, the set of $\B$-near optimal arms are defined as: 
\begin{align}
\label{eqn:near-optimal-arms}
\Barms := \left\{ a \in [K] \;\; : \;\; \sqrt{\frac{\nopt \Delta_a^2 }{2 \log \T}} \leq \B \right\}
\end{align}
We use $|\Barms|$ to denote the cardinality of $\Barms$. Our next theorem requires that there exists $\B> 0$ such that
\begin{align}
\label{eqn:large-opt-arms}
  \frac{|\Barms|}{K} \geq \alpha > 0 
  \qquad \text{for all} \;\;\;\; T \geq T_0
 \end{align}
where $T_0$ fixed. Here, the number of arms $K = K(\T)$ is allowed to grow with $T$.

\begin{theorem}
\label{thm:growing-thm}
Let Assumptions~\eqref{eqn:K-growth},~\eqref{eqn:arm-mean-diff-UB} and~\eqref{eqn:sub-Gaussian} are in force and the condition~\eqref{eqn:large-opt-arms} holds for some $\B > 0$. Then for all arms $a$: 
\begin{align}
\label{eqn:stability-large-arm}
\frac{n_{a, \T}}{\left( 1/\sqrt{n^\star} + \Delta_a/\sqrt{2 \log \T} \right)^{-2}}
\rightarrow 1 \quad \text{in probability.}
\end{align}
where $n^\star$ is the unique solution of~\eqref{eqn:n-star-eqn}.
\end{theorem}

We provide the proof of Theorem~\ref{thm:growing-thm} in Section~\ref{proof:growing-K}. A few comments on the Theorem~\ref{thm:growing-thm} are in order.

\subsubsection*{Condition on $K$:}
 The condition~\eqref{eqn:K-growth} is needed to ensure that a high probability bound inspired by the law of iterated logarithm (LIL) holds for all arm-means. The term $\log K$ arises from union bound over $K$ arms, and the term $\log \T$
comes from the UCB bonus term in~\eqref{eqn:ucb-rule-mod}. The condition~\eqref{eqn:K-growth} allows the number of arms $K$ to grow with $\T$. For instance, Theorem~\ref{thm:growing-thm} allows
\begin{align*}
K = \exp\{(\log \T)^{1 - \delta}\} \quad \text{for any} \;\; 1 > \delta > 0. 
\end{align*}

\subsubsection*{Comment on the near-optimal arm condition~\eqref{eqn:large-opt-arms}}
The condition~\eqref{eqn:large-opt-arms}
assumes that the number of \emph{near-optimal} arms are large.
The proof of the Theorem~\ref{thm:growing-thm} reveals that the
stability of the arms are related to
the stability of arm 1. In other words, if arm $1$ is stable, then all other arms are also stable. The condition~\eqref{eqn:large-opt-arms} ensures that arm $1$ is stable. This condition can be understood by looking at the characteristic equation~\eqref{eqn:n-star-eqn}. In order to infer the properties of $\nopt$, we need to make sure
that the term $\sqrt{\frac{\T}{\nopt}}$ is dominating in the term in the term $\sqrt{\T/\nopt}
+ \sqrt{\T \Delta_a^2 / 2 \log \T}$. Indeed, we have
\begin{align*}
\sqrt{\T/\nopt} \leq \sqrt{\T/\nopt}
+ \sqrt{\T \Delta_a^2 / 2 \log \T} 
&= \sqrt{\T/\nopt} 
+ \sqrt{\T/\nopt} \cdot  \sqrt{\nopt \Delta_a^2 / 2 \log \T}  \\
 &\leq (1 + \B) \cdot \sqrt{\T/\nopt} 
\end{align*}
The last inequality uses the condition~\eqref{eqn:near-optimal-arms}. There are $K$-terms in the sum~\eqref{eqn:n-star-eqn}, and since $1/K \rightarrow 0$, we need the condition~\eqref{eqn:large-opt-arms} ensures that we can recover the properties of $\nopt$ from~\eqref{eqn:n-star-eqn}. We point out that when the number of arms $K$ is finite and does not grow with $\T$, the condition~\eqref{eqn:large-opt-arms} is automatically satisfied. In particular, $\Delta_1 = 0$ by definition and 
\begin{align*}
\frac{|\Barms|}{K} \geq \frac{1}{K} > 0 \quad \text{when} \;\; K \;\; \text{is fixed}.  
\end{align*}

\subsubsection*{Condition on $\Delta_a$:}
Just like Theorem~\ref{thm:bandit-stability}, in 
Theorem~\ref{thm:growing-thm} we do not assume any specific form of the $\{\Delta_a\}_{a \in [K]}$; they can change with $\T$. The only condition that we require on $\Delta_a$ is through the condition~\eqref{eqn:large-opt-arms}.

\section{Proofs of Theorems}
\label{sec:Main-proofs}
In this section, we provide proofs of Theorems~\ref{thm:bandit-stability},~\ref{thm:CI} and~\ref{thm:growing-thm}.

\subsection{Proof of Theorem~\ref{thm:bandit-stability}}
\label{proof:thm-stability}

The bulk of the argument is based on the following concentration bound for sub-Gaussian random variables:  
\begin{lemma}
\label{lem:LIL-concentration}
Let $X_1, X_2, \ldots$ be i.i.d. 
$\lambda_a$-sub-Gaussian random variable with zero mean. Then 
\begin{align*}
\Prob\left( \exists t \geq 1 \;  : \; |\xbar_t| \geq
\lambda_a\sqrt{\frac{9}{4t} \cdot \log \frac{(\log_2 4 t)^2}{\delta}}
 \right) \leq 2\delta
\end{align*}
\end{lemma}
We prove this lemma in Section~\ref{sec:LIL-conc-proof}.  Let $\goodevent_\T$ is the following event:
\begin{subequations}
\begin{align}
\label{eqn:good-event}
\goodevent_{\T} = \left\lbrace |\xbar_{a, t} - \mu_a| \leq \lambda_a
\frac{\sqrt{7\cdot\log \log\T + 3 \log K} }{\sqrt{t}} \qquad \text{for all} \;\;
1\leq t \leq  \T, \;\; \text{and} \;\;
a \in [K]
  \right\rbrace.
\end{align}
An immediate consequence of Lemma~\ref{lem:LIL-concentration} is that 
\begin{align}
\label{eqn:good-event-prob}
\Prob(\goodevent_{\T}) \geq 1 - \frac{6}{\log \T}
\end{align}
\end{subequations}
\subsubsection{Proof idea:}
We start by proving a high probability upper and lower bound on $n_{a, T}$ on the event $\goodevent_\T$. Then stability of the arms then follows by showing that these two bounds are not too far away from each other (same asymptotic scaling). Throughout, we assume:
\begin{itemize}
\item[(a)] 
 Without loss of generality,  arm $1$ is among the optimal arms. 
\item[(b)] The high-probability good event $\goodevent_{\T}$ is in force.  We use the shorthand
\begin{align*}
\gT = \sqrt{7\log \log\T + 3 \log K}
\end{align*}
\item[(c)] We assume that $\T$ is large enough such that $\sqrt{2\log\T} \geq B \cdot g_\T$
\end{itemize}

\subsubsection{Upper bound on  $n_{a, T}$ on $\goodevent_{\T}$:}
Given any other arm $a \neq 1$, let $\lastpull{a}$ be the last time arm $a$ was pulled in $\numpull$ rounds. Then, by the UCB rule~\eqref{eqn:ucb-rule-mod} we have 
\begin{align}
\label{eqn:UCB-a-last}
\xbar_{a, \lastpull{a}}  + \bonus{a}{\lastpull{a}} \geq  \xbar_{1, \lastpull{a}}  + \bonus{1}{\lastpull{a}} 
\end{align}
Invoking~\eqref{eqn:good-event}, using the relation~\eqref{eqn:UCB-a-last} and the shorthand $\Delta_a = \mu_1 - \mu_a$ we have 
\begin{align*}
\lambda_a \lilbound{n_{a, \T_a}} + \bonus{a}{\T_a}
- \left( - \lambda_1 \lilbound{n_{1, \T_a}} + \bonus{1}{\T_a} \right) \geq \Delta_a 
\end{align*}
By inspection, 
\begin{align*}
n_{a, \T} = n_{a, \T_a} + 1 \qquad \text{and} \qquad n_{1, \T} \geq n_{1, \T_a}
\end{align*}
Combining the last two bounds we deuce
\begin{align*}
\frac{\sqrt{2 \log \T} +\lambda_a\gT}{\sqrt{n_{a, \T_a}}} \geq \frac{\sqrt{2 \log \T} -\lambda_1\gT}{\sqrt{n_{1, \T}}} 
+ \Delta_a \geq 
\frac{\sqrt{2 \log \T} -\lambda_1 \gT}{\sqrt{n_{1, \T}}} 
+ \frac{\sqrt{2 \log \T} - \lambda_1 \gT}{\sqrt{2 \log \T}} \Delta_a
\end{align*}
where the last inequality uses $\Delta_a \geq 0$. Rearranging we have 
\begin{align}
\label{eqn:armpull-a-ub}
\sqrt{\frac{2 \log \T}{n_{a, \T}}} \geq \frac{\sqrt{2 \log \T} - \lambda_1\gT}{\sqrt{2 \log \T} + \lambda_a \gT} \cdot \sqrt{\frac{n_{a, \T} - 1}{n_{a, \T}}} \cdot \left( \sqrt{\frac{2 \log \T}{n_{1, \T}}}
+ \Delta_a \right)
\end{align}

\subsubsection{Lower bound on $n_{a, \T}$ on $\goodevent_\T$:}
The lower bound for $n_{a, \T}$ follows a 
similar strategy as the upper bound proof in the last section. Let $\T_1$ be the last time arm $1$ was pulled. Then, by the UCB rule~\eqref{eqn:ucb-rule-mod} we have 
\begin{align}
\label{eqn:ucb-1-last}
\xbar_{a, \lastpull{1}}  + \bonus{a}{\lastpull{1}}
\leq 
\xbar_{1, \lastpull{1}}  + \bonus{1}{\lastpull{1}}    
\end{align}
Subtracting the population means from both sides of the relation we have that on the event $\goodevent_\T$ 
\begin{align*}
\frac{\sqrt{2 \log \T } - \lambda_a \gT}{\sqrt{n_{a, \T_1}}} \leq \frac{\sqrt{2 \log \T } +\lambda_1\gT}{\sqrt{n_{1, \T_1}}} + \Delta_a
\end{align*}
Using $n_{a, \T} \geq n_{a, \T_1}$, $n_{1, \T_1} = n_{1, \T} - 1$, and $\Delta_a\geq 0$ we conclude
\begin{align*}
\frac{\sqrt{2 \log \T } - \lambda_a \gT}{\sqrt{n_{a, \T}}} \leq \frac{\sqrt{2 \log \T } + \lambda_1 \gT}{\sqrt{n_{1, \T_1 }}} + \Delta_a \leq 
\frac{\sqrt{2 \log \T } + \lambda_1 \gT}{\sqrt{n_{1, \T} - 1}} + \frac{\sqrt{2 \log \T} + \lambda_1 \gT}{\sqrt{2 \log \T}} \Delta_a
\end{align*}
Rearranging the last bound and using $\Delta_a \geq 0$ we have 
\begin{align}
\label{eqn:armpull-a-lb}
 \sqrt{\frac{n_{1, \T} - 1}{n_{1, \T}}} \cdot \sqrt{\frac{2 \log \T }{n_{a, \T}}} \leq
\frac{\sqrt{2 \log \T} + \lambda_1 \gT}{\sqrt{2 \log \T} - \lambda_a \gT } \cdot \left( \sqrt{\frac{2 \log \T}{n_{1, \T}}}
+ \Delta_a \right) 
\end{align}

\subsubsection{Asymptotic relation between arm pulls:}
We are now ready to prove a relation between arm pull of arm $a$ with that of the optimal arms and use the relation to show the stability of arm $a$.
It follows from the bounds~\eqref{eqn:armpull-a-lb} and~\eqref{eqn:armpull-a-ub} that 
\begin{align*}
 \sqrt{\frac{n_{a, \T} - 1}{n_{a, \T}}} \cdot \frac{\sqrt{2 \log \T} -\lambda_1\gT}{\sqrt{2 \log \T} +\lambda_a\gT}  \leq \dfrac{\sqrt{\frac{2 \log \T }{n_{a, \T}}} }{\sqrt{\dfrac{2 \log \T}{n_{1, \T}}}
+ \Delta_a }  \leq \frac{\sqrt{2 \log \T} +\lambda_1 \gT}{\sqrt{2 \log \T} -\lambda_a\gT } \cdot \sqrt{\frac{n_{1, \T}}{n_{1, \T} - 1}}
\end{align*}
Squaring all sides and doing some simplification yield
\begin{align}
\label{eqn:sanwich}
 \frac{n_{1, \T} - 1}{n_{1, \T}} \cdot  \left(\frac{\sqrt{2 \log \T} -\lambda_1\gT}{\sqrt{2 \log \T} +\lambda_a\gT} \right)^2  \leq \frac{n_{a, T}}{\left( 1/\sqrt{n_{1,\T}}
+ \sqrt{\Delta_a^2 / 2 \log \T} \right)^{-2}}
\leq \left( \frac{\sqrt{2 \log \T} +\lambda_1 \gT}{\sqrt{2 \log \T} -\lambda_a \gT} \right)^2
 \cdot \frac{n_{a, \T}}{n_{a, \T} - 1}
\end{align}
For the time being let us assume:
\begin{align}
\label{eqn:large-arm-1-pull}
n_{1, \T} \geq  \frac{\T}{2K} \quad \text{on} \quad \goodevent_\T
\end{align}
This assumption along with \eqref{eqn:arm-mean-diff-UB} and the uniform boundedness of $\lambda_a$ from~\eqref{eqn:sub-Gaussian} imply that 
\begin{align*}
n_{a, \T} \rightarrow \infty \quad \text{on} \quad \goodevent_\T
\end{align*}
Combining the last observation, 
noting $\gT/\sqrt{2 \log \T} \rightarrow 0$ as $T \rightarrow \infty$ we have that for all arm $a \neq 1$ with $\Delta_a >0$ we have that   
\begin{align}
\label{eqn:asym-stability-rel}
\frac{n_{a, T}}{\left( 1/\sqrt{n_{1,\T}}
+ \sqrt{\Delta_a^2 / 2 \log \T} \right)^{-2}} \rightarrow 1.
\end{align}
The relation~\eqref{eqn:asym-stability-rel}
helps us connect the stability of every other arm with the stability of arm 1. We next proceed by doing the following:
\begin{enumerate}
\item[(a)]  Verifying assumption~\eqref{eqn:large-arm-1-pull}.
\item[(b)] The stability of arm 1. 
\item[(c)] Stability of all other arms.  
\end{enumerate}

\subsubsection{Verifying assumption~\eqref{eqn:large-arm-1-pull}}
\label{sec:arm-pull-lb}
Given any arm $a \neq 1$, let $n_{a, \T}$ be the last round arm $a$ was pulled, and the event $\goodevent_\T$ is on force. If possible, let $ n_{1, \T} \leq \frac{\T}{2K}$. Invoking the bound~\eqref{eqn:armpull-a-ub} and using $\Delta_a > 0$, we have that on the event $\goodevent_\T$
\begin{align}
\sqrt{\frac{2 \log \T}{n_{a, \T} - 1}} \geq \frac{\sqrt{2 \log \T} -\lambda_1\gT}{\sqrt{2 \log \T} + \lambda_a \gT}  \sqrt{\frac{2 \log \T}{n_{1, \T}}}
\end{align} 
By assumption, $n_{1,\T} \leq \frac{\T}{2K}$. Then, there exists $a \neq 1$ such that $n_{a, \T} \geq \T/K + 1$. 
\begin{align}
n_{1,\T} &\geq \left( \frac{\sqrt{2 \log \T} - \lambda_1 \gT}{\sqrt{2 \log \T} +\lambda_a \gT} \right)^2 
(n_{a, \T} - 1) \geq (1+o(1))\frac{\T}{\K}. 
\end{align}
Letting $\T \rightarrow \infty$ we obtain $n_{1, \T} \geq \T/K$ which is a contradiction.

\subsubsection{Stability of arm 1:}
Rearranging the~\eqref{eqn:sanwich} relation we have 
\begin{align}
1 = \sum_{a} \frac{n_{a, \T}}{\T} 
\geq \frac{n_{1, \T} - 1}{n_{1, \T}} \sum_{a}
\left( \frac{\sqrt{2 \log \T} - \lambda_1\gT}{\sqrt{2 \log \T} +\lambda_a\gT} \right)^2 \cdot  
\frac{\left( 1/\sqrt{n_{1,\T}}
+ \sqrt{\Delta_a^2 / 2 \log \T} \right)^{-2}}{\T}
\end{align}
Again from~\eqref{eqn:sanwich} we also have 
\begin{align}
 \frac{\T - K}{\T} =  \sum_{a} \frac{n_{a, \T} - 1}{\T} 
\leq \sum_{a}
\left( \frac{\sqrt{2 \log \T} +\lambda_1\gT}{\sqrt{2 \log \T} -\lambda_a \gT} \right)^2 \cdot
\frac{\left( 1/\sqrt{n_{1,\T}}
+ \sqrt{\Delta_a^2 / 2 \log \T} \right)^{-2}}{\T}. 
\end{align}
Combining the last two bounds and using $n_{1,\T} \geq \T/2K$ from~\eqref{eqn:large-arm-1-pull} we have 
%
%
\begin{align}
&\sum_{a } 
\frac{1}{\left( \sqrt{\T/n_{1,\T}}
+ \sqrt{\T \Delta_a^2 / 2 \log \T} \right)^{2}} \rightarrow 1 
\ \ \text{as $\T\to\infty$.}
\label{eqn:convergence-to-one}
\end{align}
From here, we would like to show that $n_{1,\T}/\T$ converges to some \emph{non-random} quantity asymptotically. For a given $\T$, let $n^\star = n^\star (\T)$ be the unique solution to the following:
\begin{align}
 \sum_{a } 
\frac{1}{\left(\sqrt{\T/\nopt}
+ \sqrt{\T \Delta_a^2 / 2 \log \T} \right)^{2}} = 1
\end{align}
To see such a solution always exists and unique, given $\T$ and $\{\Delta_a\}_{a > 1}$ define the function $f: \mathbb{R}_{+} \mapsto \mathbb{R}$
\begin{align*}
f(y) = \sum_{a } 
\frac{1}{\left(\sqrt{\T/y}
+ \sqrt{\T \Delta_a^2 / 2 \log \T} \right)^{2}} - 1
\end{align*}
By construction, the function $f$ is strictly monotone in $y \in (0, \infty)$. 
It is easy to see using the non-negativity of $\Delta_a's$ that $f(\T/k) \leq 0$, and $f(\T) > 0$ (assuming $K > 1$). This implies that for all $\T \geq K$ 
\begin{align}
\label{eqn:nopt-bound}
\frac{1}{K} \leq \frac{\nopt}{\T} \leq 1. 
\end{align}   
We are now ready to prove $\frac{n_{1,\T}}{\nopt}  \rightarrow 1$. We prove this via contradiction. 
Suppose there exists  $\epsilon > 0$ such that
\begin{align}
\label{eqn:non-convergence}
n_{1,\T}/\T > \nopt/\T + \epsilon
\qquad \text{for infinitely many} \;\; \T.
\end{align}
From the equation~\eqref{eqn:convergence-to-one} we have that there exists $T_\epsilon$ such that 
\begin{align}
\label{eqn:hypothesis}
1 - \frac{\epsilon}{2} \leq  \sum_{a} 
\frac{1}{\left( \sqrt{\T/n_{1,\T}}
+ \sqrt{\T \Delta_a^2 / 2 \log \T} \right)^{2}} \leq 1 + \frac{\epsilon}{2}
\end{align}
for all $\T \geq \T_\epsilon$.
By assertion~\eqref{eqn:non-convergence}, there exists $\Tprime \geq \T_\epsilon$ such that $n_{1,\Tprime}/\Tprime > \nopt/\Tprime + \epsilon$. We then have that for all $\Delta_a \geq 0$ 
\begin{align*}
 \frac{1}{\left(\sqrt{\Tprime/n_{1, \Tprime}}
+ \sqrt{ \Delta_a^2 / 2 \log \Tprime} \right)^{2}} \geq \frac{1}{\left(\sqrt{\Tprime/\nopt}
+ \sqrt{\Tprime \Delta_a^2 / 2 \log \Tprime} \right)^{2}}.
\end{align*}
Combining the last two relations we have 
\begin{align*}
 \frac{n_{1,\Tprime}}{\Tprime} + \sum_{a \neq 1}  \frac{1}{\left(\sqrt{\Tprime/n_{1, \Tprime}}
+ \sqrt{ \Delta_a^2 / 2 \log \Tprime} \right)^{2}} &\geq  \frac{\nopt}{\Tprime} + \sum_{a \neq 1} \frac{1}{\left(\sqrt{\Tprime/\nopt}
+ \sqrt{\Tprime \Delta_a^2 / 2 \log \Tprime} \right)^{2}} + \epsilon \\
&= 1 + \epsilon. 
\end{align*}
This is a contradiction with the hypothesis~\eqref{eqn:hypothesis}. A similar argument holds when $n_{1,\T}/\T < \nopt/\T - \epsilon$ for infinitely many $\T's$. Thus we conclude  
\begin{align*}
n_{1,\T}/\T - \nopt/\T \rightarrow 0
\end{align*}
Combining this fact that with~\eqref{eqn:nopt-bound} we have 
\begin{align}
\label{eqn:stability-optimal-arm}
\frac{n_{1,\T}}{\nopt}  \rightarrow 1
\end{align}
This proves the stability of arm 1 by noting that $\Prob(\goodevent_{\T}) \geq 1 - \frac{6}{\log \T}$. 

\subsubsection{Stability of all other arms:}
Fix any arm $a \neq 1$. From equation~\eqref{eqn:sanwich} it suffices to show that on $\goodevent_\T$
\begin{align*}
\frac{\left( 1/\sqrt{n^\star}
+ \sqrt{\Delta_a^2 / 2 \log \T} \right)^{2}}{\left( 1/\sqrt{n_{1,\T}}
+ \sqrt{\Delta_a^2 / 2 \log \T} \right)^{2}}
\rightarrow 1
\end{align*}
We have 
\begin{align*}
\frac{\left( 1/\sqrt{n^\star}
+ \sqrt{\Delta_a^2 / 2 \log \T} \right)}{\left( 1/\sqrt{n_{1,\T}}
+ \sqrt{\Delta_a^2 / 2 \log \T} \right)} 
& = \frac{\left( 1
+ \sqrt{n^\star\Delta_a^2 / 2 \log \T} \right)}{\left( \sqrt{\frac{n^\star}{n_{1,\T}}}
+ \sqrt{ n^\star \Delta_a^2 / 2 \log \T} \right)} \\
& = 1 +  \frac{\left( \sqrt{\frac{n^\star}{n_{1,\T}}} - 1 \right)}{\left( \sqrt{\frac{n^\star}{n_{1,\T}}}
+ \sqrt{ n^\star \Delta_a^2 / 2 \log \T} \right)} \rightarrow 1. 
\end{align*}
The last line follows from the fact that for large $\T$ we have $1/2 \leq \sqrt{\frac{n^\star}{n_{1,\T}}} \leq 2$, and $\frac{n^\star}{n_{1,\T}} \rightarrow 1$. 
This completes the proof of Theorem~\ref{thm:bandit-stability} by noting that $\Prob(\goodevent_{\T}) \geq 1 - \frac{6}{\log \T}$. 

\subsection{Proof of Theorem~\ref{thm:CI}}
The proof of this theorem utilizes Lai and Wei~\cite[Theorem 3]{lai1982least} and Theorem~\ref{thm:bandit-stability}. Indeed,
writing the multiarmed bandit problem as a stochastic regression model from Lai and Wei~\cite{lai1982least} we see that Theorem~\ref{thm:bandit-stability} ensures that the covariate stability condition~\cite[Condition 4.2]{lai1982least} holds. Additionally, the condition 
~\cite[Condition 4.3]{lai1982least} holds since $\nopt \rightarrow \infty$ and $\Delta_a/\sqrt{2 \log \T} = o(1)$ (see for instance~\eqref{eqn:arm-a-stability}). Thus, invoking~\cite[Theorem 3]{lai1982least} we have
\begin{align}
\label{eqn:joint-normality}
\left(\frac{\xbar_{1,T} - \mu_1}{\sqrt{n_{1,\T}}}, \ldots, \frac{\xbar_{K,T} - \mu_k}{\sqrt{n_{K,\T}}} \right) \stackrel{d}{\rightarrow} \Ncal\left(0, \mathbf{\Sigma} \right). 
\end{align}
Here, 
\begin{align}
\Sigma = \textrm{diag} \left( \sigma^2_1, \ldots, \sigma^2_K \right).
\end{align}
Now given any consistent estimate $\rvarhat_a$ of $\rvar_a$ we combine~\eqref{eqn:joint-normality} with Slutsky's theorem to produce
\begin{align*}
\left(\frac{\sigmahat_{1,\T} \cdot (\xbar_{1,T} - \mu_1)}{\sqrt{n_{1,\T}}}, \ldots, \frac{\sigmahat_{K,\T} \cdot (\xbar_{K,T} - \mu_k)}{\sqrt{n_{K,\T}}} \right) \stackrel{d}{\rightarrow} \Ncal\left(0, \mathbf{\Sigma} \right)
\end{align*}
The coverage claim of Theorem~\ref{thm:CI} is now immediate. This completes the proof of Theorem~\ref{thm:CI}.

\subsection{Proof of Theorem~\ref{thm:growing-thm}}
\label{proof:growing-K}
Using~\eqref{eqn:K-growth} and following the proof of Theorem~\ref{thm:bandit-stability} we have that on the event $\goodevent_{\T}$ 
\begin{subequations}
\begin{align}
\frac{n_{a, T}}{\left( 1/\sqrt{n_{1,\T}}
+ \sqrt{\Delta_a^2 / 2 \log \T} \right)^{-2}} &\rightarrow 1 \quad \text{and} \quad 
\label{eqn:arm-a-cond}
  \\
\sum_{a } 
\frac{1}{\left( \sqrt{\T/n_{1,\T}}
+ \sqrt{\T \Delta_a^2 / 2 \log \T} \right)^{2}} &\rightarrow 1. 
\label{eqn:sum-arm-cond}
\end{align}
\end{subequations}

We now first show that $\frac{n_{1, \T}}{n^\star} \rightarrow 1$ in probability; meaning that arm $1$ is stable. To this end, we define 
\begin{align*}
f(y) = \sum_{a } 
\frac{1}{\left(\sqrt{\T/y}
+ \sqrt{\T \Delta_a^2 / 2 \log \T} \right)^{2}} 
\end{align*}
Recall that $f(y)$ is monotonically increasing in $y$, and $f(n^\star) = 1$. Additionally, using the bounds~\eqref{eqn:nopt-bound},~\eqref{eqn:sanwich} and~calculations from Section~\ref{sec:arm-pull-lb} we have that on $\goodevent_\T$
\begin{align*}
n^\star \geq \frac{\T}{K} \rightarrow \infty \quad \text{and} \quad n_{a, \T} \rightarrow \infty \quad \text{for all arms} \;\; a.  
\end{align*}

\subsubsection{Stability of arm 1:} 
Given any $1 > \epsilon > 0$, and let $\T_\epsilon$ be such that $n_{1, \T} \geq \nopt (1 + \epsilon)$. We have 
\begin{align*}
f(\nopt(1 + \epsilon)) &= 
\sum_{a \in \Barms}  
\frac{1}{\left(\sqrt{\T/\nopt(1 + \epsilon)}
+ \sqrt{\T \Delta_a^2 / 2 \log \T} \right)^{2}} + \sum_{a \notin \Barms } 
\frac{1}{\left(\sqrt{\T/\nopt(1 + \epsilon)}
+ \sqrt{\T \Delta_a^2 / 2 \log \T} \right)^{2}} \\
& \stackrel{(i)}{\geq} \sum_{a \in \Barms}  
\frac{1}{\left(\sqrt{\T/\nopt(1 + \epsilon)}
+ \sqrt{\T \Delta_a^2 / 2 \log \T} \right)^{2}} + \sum_{a \notin \Barms } 
\frac{1}{\left(\sqrt{\T/\nopt}
+ \sqrt{\T \Delta_a^2 / 2 \log \T} \right)^{2}} \\
&=  \left( \sum_{a \in \Barms}  
\frac{1}{\left(\sqrt{\T/\nopt(1 + \epsilon)}
+ \sqrt{\T \Delta_a^2 / 2 \log \T} \right)^{2}} - \sum_{a \in \Barms}  
\frac{1}{\left(\sqrt{\T/\nopt}
+ \sqrt{\T \Delta_a^2 / 2 \log \T} \right)^{2}} \right) \\
& \qquad + \sum_{a \in \Barms}  
\frac{1}{\left(\sqrt{\T/\nopt}
+ \sqrt{\T \Delta_a^2 / 2 \log \T} \right)^{2}} + \sum_{a \notin \Barms } 
\frac{1}{\left(\sqrt{\T/\nopt}
+ \sqrt{\T \Delta_a^2 / 2 \log \T} \right)^{2}} \\
& \stackrel{(ii)}{=} 1 +  \left( \sum_{a \in \Barms}  
\frac{1}{\left(\sqrt{\T/\nopt(1 + \epsilon)}
+ \sqrt{\T \Delta_a^2 / 2 \log \T} \right)^{2}} - \sum_{a \in \Barms}  
\frac{1}{\left(\sqrt{\T/\nopt}
+ \sqrt{\T \Delta_a^2 / 2 \log \T} \right)^{2}} \right) 
\end{align*} 
The line $(i)$ above uses $\epsilon > 0$ and $\Delta_a \geq 0$, and the equality $(ii)$ uses the fact that $f(\nopt) = 1$. Doing some simplification using $1 > \epsilon > 0$, and the definition of~\eqref{eqn:near-optimal-arms} we have 
\begin{align*}
\sum_{a \in \Barms}  
\frac{1}{\left(\sqrt{\T/\nopt(1 + \epsilon)}
+ \sqrt{\T \Delta_a^2 / 2 \log \T} \right)^{2}} - \sum_{a \in \Barms}  
\frac{1}{\left(\sqrt{\T/\nopt}
+ \sqrt{\T \Delta_a^2 / 2 \log \T} \right)^{2}}  \geq \frac{\epsilon \nopt |\Barms|}{\T (1 + 2 \B)^2(1 + \B)^2}
\end{align*}
Putting together the pieces
\begin{align*}
f(\nopt(1 + \epsilon)) &\geq 1 +  \frac{ |\Barms| \nopt}{\T} \cdot \frac{\epsilon}{(1 + 2 \B)^2 (1 + \B)^2} \\
& \geq \frac{ |\Barms| }{K} \cdot \frac{\epsilon}{(1 + 2 \B)^2 (1 + \B)^2} \\
& \geq \frac{\alpha \epsilon}{(1 + 2 \B)^2 (1 + \B)^2}
\end{align*}
The second inequality above uses the fact that $\nopt \geq \T/K$, and the third inequality above uses the assumption~\eqref{eqn:large-opt-arms}.
Putting together the pieces we have 
\begin{align*}
f(n_{1, \T}) \geq f(\nopt(1 + \epsilon)) \geq 1 + \frac{\alpha \epsilon}{(1 + 2 \B)^2 (1 + \B)^2}
\end{align*}
Using a similar argument for $n_{1, \T} \leq \nopt(1 - \epsilon)$ we have 
\begin{align*}
f(n_{1, \T}) \leq f(\nopt(1 - \epsilon)) \leq 1  -  \frac{\alpha \epsilon}{(1 + 2 \B)^2 (1 + \B)^2}  
\end{align*}
Since $\epsilon > 0$ was arbitrary, $B > 0$ is a constant, and $f(n_{1, \T}) \rightarrow 1$ we conclude that $\frac{n_{1, \T}}{\nopt} \rightarrow 1$. This completes the stability proof of arm $1$. We now move on to the stability proof of all other arms. 

\subsubsection{Stability of all arms}
Fix any arm $a \neq 1$, and the event $\goodevent_\T$ is in force. Using~\eqref{eqn:arm-a-cond} it suffices to show that 
\begin{align*}
\frac{\left( 1/\sqrt{n^\star}
+ \sqrt{\Delta_a^2 / 2 \log \T} \right)^{2}}{\left( 1/\sqrt{n_{1,\T}}
+ \sqrt{\Delta_a^2 / 2 \log \T} \right)^{2}}
\rightarrow 1
\end{align*}
We have 
\begin{align*}
\frac{\left( 1/\sqrt{n^\star}
+ \sqrt{\Delta_a^2 / 2 \log \T} \right)}{\left( 1/\sqrt{n_{1,\T}}
+ \sqrt{\Delta_a^2 / 2 \log \T} \right)} 
& = \frac{\left( 1
+ \sqrt{n^\star\Delta_a^2 / 2 \log \T} \right)}{\left( \sqrt{\frac{n^\star}{n_{1,\T}}}
+ \sqrt{ n^\star \Delta_a^2 / 2 \log \T} \right)} \\
& = 1 +  \frac{\left( \sqrt{\frac{n^\star}{n_{1,\T}}} - 1 \right)}{\left( \sqrt{\frac{n^\star}{n_{1,\T}}}
+ \sqrt{ n^\star \Delta_a^2 / 2 \log \T} \right)} \rightarrow 1. 
\end{align*}
The last line follows from the fact that for large $\T$ we have $1/2 \leq \sqrt{\frac{n^\star}{n_{1,\T}}} \leq 2$, and $\frac{n^\star}{n_{1,\T}} \rightarrow 1$. 
This completes the proof of Theorem~\ref{thm:bandit-stability} by noting that $\Prob(\goodevent_{\T}) \geq 1 - \frac{6}{\log \T}$. This completes the proof of Theorem~\ref{thm:growing-thm}.

\section{Discussion}
\label{sec:Discussion}
This paper establishes a novel stability property of the upper confidence bound (UCB) algorithm in the context of multi-armed bandit problems. This property makes the downstream statistical inference straightforward; for instance, classical statistical estimators remain asymptotically normal even though the data collected is not iid. Moreover, we show that when the number of arms $K$ are large, and potentially allowed to grow with the number of total arm pulls. Finally,  our result imply that the UCB algorithm is \emph{fair}. Concretely, if two arm means are close, the UCB algorithm will select both arms equal number of times in the long run (asymptotically).

While these findings represent a significant advance, several open questions remain. Future research could explore the extension of these results to heavy-tailed distributions, investigate stability properties of other popular bandit algorithms, and examine potential trade-offs between stability and regret minimization. Put differently, it would be interesting to see if it is possible to verify the stability of other popular bandit algorithms or propose a stable analogues without increasing the regret by significant amount. We believe establishing stability properties of reinforcement learning algorithms would improve the reliability and reproducibility of reinforcement learning systems in practice.

\section*{Acknowledgments}
This work was partially supported by the National Science Foundation Grant DMS-2311304 to KK, and the National Science Foundation Grants CCF-1934924, DMS-2052949 and DMS-2210850 to CHZ.

\bibliographystyle{imsart-number} 
\bibliography{ref}

\appendix 

\section{Proofs of Corollaries}
In this section, we prove Corollary~\ref{cor:var-estimate} on the consistency of the variance estimator $\widehat{\sigma}^2_a$. 
\newcommand{\subGauss}{\lambda}

\subsection{Proof of Corollary~\ref{cor:var-estimate}}
\label{proof-cor-uneq-var}
Fix an arm $a \in [K]$. We have 
\begin{align*}
\widehat{\sigma}^2_a = \frac{1}{n_{a, \T}} \sum_{t = 1}^\T X_t^2 \cdot 1_{\{A_t = a\} } - \left(\frac{1}{n_{a, \T}} \sum_{t = 1}^\T X_t \cdot 1_{\{A_t = a\} } \right)^2
\end{align*}
It suffices to show that 
\begin{align}
\label{eqn:consistency-claim}
\frac{1}{n_{a, \T}} \sum_{t = 1}^\T X_t \cdot 1_{\{A_t = a\} }\stackrel{p}{\rightarrow} \mu_a \quad \text{and} \quad 
\frac{1}{n_{a, \T}} \sum_{t = 1}^\T X_t^2 \cdot 1_{\{A_t = a\} } \stackrel{p}{\rightarrow} \mu_a^2 + \sigma_a^2
\end{align}
Finally, Theorem~\ref{thm:bandit-stability} along with assumption~\eqref{eqn:arm-mean-diff-UB} ensures that 
\begin{align*}
n_{a, \T} \stackrel{p}{\rightarrow} \infty \quad \text{as} \;\; \T \rightarrow \infty. 
\end{align*}
Now the claim~\eqref{eqn:consistency-claim} follows from~\cite[Proposition 3.2]{shin2019bias} on the consistency of sample average; see also~\cite[Theorem 2.1]{gut2009stopped}. This completes the proof.

\section{Proof of Technical Lemmas}
In this section we prove a few technical results that are used in the proof the theorems and corollaries. 
\subsection{Proof of Lemma~\ref{lem:LIL-concentration}}
\label{sec:LIL-conc-proof}
The proof is based on the work of~\cite{balsubramani2014sharp}, and an unpublished \href{http://blog.wouterkoolen.info/QnD_LIL/post.html}{ article}~\footnote{\url{http://blog.wouterkoolen.info/QnD_LIL/post.html}} by Wouter M. Koolen. We use this with minimal modification. The proof utilizes Doob's inequality which states that for any super-Martingale $\{Z_t\}_{t \geq 0}$ with $Z_0 = 1$ we have 
\begin{align*}
\Prob\left(\exists \; t \; : \; Z_t \geq \frac{1}{\delta}\right) \leq \delta
\end{align*}
Let $S_t = \sum_{j \leq t} X_j$. Define $Z_0 = 1$ and $Z_t$ as   
\begin{align*}
Z_t = \sum_{i = 1}^\infty \gamma_i \exp\left\{ \eta_i S_t - \frac{t \eta_i^2}{2} \right\}
\end{align*}
we will choose $\eta_i \geq 0$ and $\gamma_i \geq 0$ shortly. But let us first show that $Z_t$ is a super-Martingale. Let $\filt{t}$ denote the natural filtration generated by the sequence $\{X_t\}_{t \geq 1}$

\begin{align*}
\Exs[ Z_{t + 1} \mid \filt{t} ]
&= \sum_{i = 1}^{\infty} \gamma_i \exp \left\{ \eta_i S_{t} - \frac{t\eta_i^2}{2}  \right\} \Exs \left\{ \eta_i Z_t   - \frac{\eta_i^2}{2} \mid \filt{t} \right\} \notag \\
&\leq \sum_{i = 1}^{\infty} \gamma_i \exp \left\{ \eta_i S_{t} - \frac{t\eta_i^2}{2}  \right\} = Z_t
\end{align*}
where the first inequality above follows from the $1$-sub Gaussian property of the random variable $Z_t$. Given a $\delta > 0$, we now set 
\begin{align*}
\gamma_i = \frac{1}{i(i + 1)} \;\; \text{and}
\;\; \eta_i = \sqrt{\frac{2 \log \frac{1}{\delta \gamma_i}}{2^i}}  
\end{align*}
It now remains to understand the event $\left\{ Z_t \geq 1/\delta \right\}$. First note that 
\begin{align*}
\left\{ Z_t \geq 1/\delta \right\}
& \supseteq  \left\{ \max_i \gamma_i \exp \left\{ \eta_i S_{t} - \frac{(t)\eta_i^2}{2} \right\} \geq 1/\delta \right\} \\
& = \left\{ \frac{S_t}{t} \geq \min_i \left( \sqrt{\frac{t}{2^i}} + \sqrt{\frac{2^i}{t}} \right) \sqrt{\frac{\log \frac{i(i + 1)}{\delta} }{2t}} \right\} \\
&\supseteq \left\{ \bar{X}_t \geq \sqrt{\frac{9\log \frac{(\log 4t)^2}{\delta} }{4t}}  \right\} 
\end{align*}
where the last line follows by taking $i = \lceil{\log_2 t\rceil}$.  This completes the proof.

\end{document}